# STORM - A Novel Information Fusion and Cluster Interpretation Technique

Jan Feyereisl and Uwe Aickelin

School of Computer Science, The University of Nottingham, NG8 1BB, UK
{jqf,uxa}@cs.nott.ac.uk

**Abstract.** Analysis of data without labels is commonly subject to scrutiny by unsupervised machine learning techniques. Such techniques provide more meaningful representations, useful for better understanding of a problem at hand, than by looking only at the data itself. Although abundant expert knowledge exists in many areas where unlabelled data is examined, such knowledge is rarely incorporated into automatic analysis. Incorporation of expert knowledge is frequently a matter of combining multiple data sources from disparate hypothetical spaces. In cases where such spaces belong to different data types, this task becomes even more challenging. In this paper we present a novel immune-inspired method that enables the fusion of such disparate types of data for a specific set of problems. We show that our method provides a better visual understanding of one hypothetical space with the help of data from another hypothetical space. We believe that our model has implications for the field of exploratory data analysis and knowledge discovery.

## 1 Introduction

The machine learning community embraces two types of learning that encompass the majority of algorithms present within this field. Supervised learning, where examples of data of interest exist, and unsupervised learning where no explicit examples are available. When examples are present, a decision function can be found by exploiting the knowledge of such examples. On the other hand, without such knowledge, only similarity between data can be exploited in order to find groups of data that share some common attributes [1].

The human immune system has inspired a number of algorithms that fall into these two categories [3], yet it does not simply operate only within these two realms. Knowledge is embedded within DNA passed down from generation to generation, eventually transforming into biological entities or functionalities that provide additional knowledge to what is learned during the lifetime of a living being. One example of such inherited knowledge can be found within Toll-like receptors (TLR), present on several types of immune cells [4].

In this work we will show that an analogy of TLRs provides an insight into a third class of learning that encodes knowledge that is not within the same hypothetical space as knowledge encoded within a training or testing dataset. Such a type of learning becomes especially useful where no labelled examples exist, but some knowledge about classes of interest is acknowledged. We believe that such incorporation provides for better understanding of underlying data based on more than blind function approximation.

In the remaining sections of this paper we first outline the functionality of TLRs, followed by our hypothesis. A description of the underlying machine learning algorithm is then presented. A theoretical specification of our StOrM model is then described, outlining a cluster interpretation technique stemming from our model. This is followed by experimental evidence confirming our hypothesis.

## 2  Toll-Like Receptors

TLRs are a set of receptors on the surface of immune cells which act as sensors to foreign microbial products. One interesting aspect of these receptors is that they act like piano keys. A different sound is played when a different key or combination of keys is pressed at once. In a similar way, when a single receptor senses a chemical, it results in a different action performed than when a number of receptors sense various chemicals within a specific time period [7].

A simple definition of TLRs is that they are the initial detectors of pathogens attacking a system. They sound an alarm when they encounter certain virus- or bacteria-specific chemicals, which trigger a cascade of events potentially resulting in an immune response. Unlike in many other parts of the immune system, all this is possible due to evolved knowledge passed down from parents to offspring over many generations. For a detailed description of research in the area of TLRs the reader is directed to [4].

### 2.1  TLRs and Learning

The idea of TLRs for the purpose of learning is a very simple one. It is a direct translation of the receptors' functionality within the human body. TLR is a signature or a function, encoding some known truth. A set of TLRs on the other hand encode a class of interest. Our hypothesis is that by formulating data from one hypothetical space as a set of TLRs, we will be able to combine information from two disparate spaces in a way that will give us a better understanding of the problem. This fusion of information is especially useful in cases where some knowledge about data of interest is known, yet limited amount or no examples exist.

Vapnik [10] also realised this missing area between supervised and unsupervised learning and proposed a related idea, which he terms "master-class learning". In his work however, he proposes an extension of a supervised learning setting, where a training dataset belonging to space $\chi$, with labels, is supplemented with an additional description of this data in another space $\chi^*$. This description of data is called "hidden information", which can exist in the form of expert knowledge describing the underlying problem. Vapnik combines his model with his support vector machine algorithm and shows that a poetic description [10] of a set of images of numbers provides more useful knowledge for learning than a higher resolution image, which holds more "technical" information about the underlying digits. In Vapnik's work a poetic description is a poet's textual depiction of the underlying image. Vapnik's aim is to improve the classification performance of supervised function estimation based on this "hidden information". In contrast, we propose a method to fuse expert knowledge (one hypothetical space) with "technical" information (another hypothetical space) for the purpose of unsupervised analysis and visualisation for better exploratory data analysis.

## 2.2 TLR Model

Using Vapnik's notation we can formalise our TLR analogy. In supervised learning a pair $(x_i, c_i)$ is given, where $x_i$ denotes a vector of some dimensionality and $c_i$ denotes a class label. In unsupervised learning only $(x_i)$ is given. In our model a tuple $(x_i, s_i)$ is given, where $s_i$ denotes a data structure which encodes some additional knowledge or side information about a data instance. As our knowledge might be limited, $s_i$ might be empty when no knowledge exists. Data in $s_i$ belongs to a space $\chi^*$, which is related to $\chi$. By this we mean that there exists a meaningful correlation between $\chi^*$ and $\chi$. Without such correlation, we can state that the knowledge represented in $\chi^*$ is not descriptive of $\chi$. In order to incorporate expert knowledge as part of a learning mechanism, we propose the use of an established machine learning technique, which provides numerous features that are beneficial to our model. Description of this technique follows.

## 3 The Self-Organising Map

Self-organising network algorithms provide a number of mechanisms desirable for many computational tasks. Features such as manifold learning, dimensionality reduction, multidimensional scaling as well clustering and visualisation through self-organisation are all mechanisms that in combination provide a suitable basis for the incorporation of our TLR analogy and for its better understanding. One type of self-organising network is the Self-Organising Map (SOM) algorithm developed by Teuvo Kohonen [5]. For a detailed description of the algorithm the reader is referred to Kohonen's extensive book on this topic [6]. It is important to note however that the SOM is only one of many types of algorithms that we believe our model can be applied to. In general any topology preserving and manifold learning algorithm could possibly be extended in order to achieve a comparable outcome. SOM was chosen due to its simplicity, speed and visualisation capabilities.

## 4 StOrM - TLR Enhanced SOM

### 4.1 Model Overview

In order to fuse knowledge from disparate hypothetical spaces with the unsupervised learning outcome of the SOM, we extend the original algorithm in a number of ways. These can be divided into two categories. Firstly data fusion and correlation is performed through an extension of the original SOM. Secondly a cluster interpretation algorithm is devised, exploiting the extended SOM in order to provide a better visual representation of underlying data.

Fusion of Hypothetical Spaces - In the original SOM, the algorithm is presented with an input

$$x = [\xi_1, \xi_2, \ldots, \xi_n]^T \in \Re^n \tag{1}$$

where $\xi$ is an attribute of the input vector x. This is the "technical" information on which the SOM is trained. In our experiments this vector, for example, comprises normalised real-valued data describing a time-based snapshot of behaviour of one running process according to a number of host based measures. In our model two additional inputs exists. First, input for the separate hypothetical space, which is in the form

$$s = [\varsigma_1, \varsigma_2, \ldots, \varsigma_m]^T \tag{2}$$

where s is a vector, however this time comprising of an arbitrary number, m, of variables $\varsigma$ that encode instance specific information related to x, from space $\chi^*$. In our experiments this vector comprises all API calls that the process, whose snapshot is encoded by x, imports. Second, a vector e encoding our expert knowledge exists,

$$e = [\epsilon_1, \epsilon_2, \ldots, \epsilon_m]^T \tag{3}$$

Here vector e can however comprise not only of fixed data, $\epsilon$, but also of functions, $\epsilon$, that express the expert's set of knowledge that is desired to be observed and identified within s or x. In contrast to s, vector e is a global, rather than a per instance vector. Returning to our immunological analogy, e is our repertoire of TLR receptors, where each $\epsilon$ is an individual receptor. In our experiments e is a vector of strings that is representative of the majority of API call names associated with networking functionality in the Windows OS. Each element $\epsilon$ is representative of a subclass of networking functions (e.g. $\epsilon_1$ = 'http').

Once the original SOM is presented with input x, it finds the most similar prototype vector and its associated node, also called the "winner node" c,

$$\|x - m_c\| = \min_i \{\|x - m_i\|\} \tag{4}$$

This node along with all nodes in its immediate neighbourhood is subsequently subject to a learning process, over a predefined time period, with a discrete time-coordinate t,

$$m_i(t + 1) = m_i + h_{ci}(t)[x(t) - m_i(t)] \tag{5}$$

Here the function $h_{ci}$ denotes the neighbourhood function which determines the amount by which a prototype vector $m_i$ is affected during the learning process. This depends on node i's distance from the "winner node" c. Generally, the following smoothing kernel, written in terms of the Gaussian function, is used,

$$h_{ci}(t) = \alpha(t) \cdot exp\left(-\frac{\|r_c - r_i\|^2}{2\sigma^2(t)}\right) \qquad (6)$$

where α denotes the learning rate and σ defines the width of the kernel. Variables $r_c$ and $r_i$ are location vectors of the winner node c and currently observed node i in the output grid. For more detailed explanation see Kohonen's book [6].

Once this learning terminates, the algorithm presents a discrete regular grid containing a lower dimensional representation of the input which preserves topology of the learned data. This grid comprises of nodes i, with which reference vectors $m_i$ are associated, that hold the learned information. In our model an additional reference vector, $l_i$, exists. This vector learns information according to the following additional computation step,

$$l_c(t+1) = l_c(t) \vee \Lambda(s(t), e(t), x(t)) \qquad (7)$$

where Λ is a matching function that evaluates which elements of s and x satisfy conditions specified within e. In other words this function evaluates which TLRs have been activated for the currently observed winner node c. Operator $\vee$ is a Boolean operator on elements of l and the output of Λ, i.e. $l_c$ learns which known truth has been observed by a winner node c over the duration of the SOM learning process. The result of this additional step is that desired knowledge from the separate hypothetical space is correlated with the produced map. This correlation is exhibited by the enhanced SOM output containing nodes i which have two associated reference vectors. One "technical", $m_i$, from $\chi$ and one of correlated expert knowledge, $l_i$, extracted from the separate hypothetical space $\chi^*$. In our experiments $l_i$ learns whether node i has ever been deemed a winner for some input x associated with a process that uses Windows networking API functions. The encoding of information from $\chi^*$ can now be used, for example, for cluster interpretation and labelling. In our experiments exploited to delineate a cluster of nodes responsible for networking behaviour. It is important to note that this information can however be exploited in other ways to enhance the output of the SOM. For example by affecting the actual SOM learning function, in order to include information from $\chi^*$ in the actual map generation process.

**Cluster Interpretation -** Due to topology preserving nature of the SOM, newly introduced expert knowledge can now be used to identify clusters of interest within the output map. Our proposed cluster interpretation technique comprises of two steps. Firstly an established algorithm called the Unified Distance Matrix (U-Matrix) [9] is exploited in order to find nodes which possibly lie on cluster boundaries. This information is subsequently used by a step which connects nodes with similar TLR information, $l_i$, in order to delineate clusters of interest.

Cluster Boundary Search - The cluster boundary search algorithm exploits an idea incorporated within the U-Matrix visualisation technique. This method shows dissimilarities between neighbouring nodes in order to highlight where possible cluster boundaries lie. In order to find nodes which lie on such boundaries, we propose to collect information about all inter-node distances along both i and j dimensions. Once this information is obtained, it is subjected to a variability function which identifies distances between nodes that significantly differ from distances between the majority of nodes on the map. This function is subject to further future research, however here we provide some pointers on how it might operate and how it is employed in our experiments. A well trained SOM map can be thought of as comprising of clusters existing within a dataset on which it is trained. A careful examination of proportion of clusters versus inter-cluster nodes needs to be performed in order to determine such proportion correctly. An example of a quantitative measure that could be used to represent such ratio can be found as follows. Assuming 25% of nodes within a generated map are inter-cluster nodes, we define any inter-node distance above the 3rd quantile of all inter-node distances as lying on a cluster boundary. Thus we can label all nodes whose dissimilarity is above the 3rd quantile threshold, as being a cluster boundary node.

Labelling using Node Connectivity - Once we obtain cluster boundary information, we can use our learned TLR knowledge in order to label clusters that exist within the SOM map. In order to achieve this, the preservation of topology within a SOM map is exploited by exploring neighbouring nodes within a segment of a map, delineated by boundary nodes. The labelling algorithm traverses all nodes $m_i$ within our map and connects nodes which lie within a neighbouring region. A region is usually bounded by the previously found boundary nodes. Once all possible nodes are connected, a connected region is evaluated for the most frequently occurring activated TLR type. This information then provides a label for all nodes within such connected region. An example map where the result of these steps can be seen is in Fig. 1(c).

In the literature other SOM cluster interpretation techniques exist. These techniques exploit various additional machine learning methods to achieve their goal. For example a two-stage procedure, where SOM output is fed into a traditional clustering technique, such as k-means [11], or hierachical clustering [12], evaluated with the help of numerous cluster validity indices. Similarly to our work Brugger et al. [2] also exploit the topographic surface of the SOM, however with the help of an algorithm called clusot, rather than the U-Matrix method used in our work.

It is important to note here that our model is not only a cluster interpretation or labelling technique. Our model provides a method for correlating data from disparate sources, which in this paper is used to identify and subsequently label clusters of interest, without traditionally labelled data. The model can however be used for other purposes which could benefit from the exploitation of the data fusion that is the result of our TLR functionality. As mentioned before, for example, the SOM learning function can be affected to take into account data from the separate hypothetical space. This is one of our future research goals.

## 5 Experiments

Two experiments were performed in order to validate our proposed model. One to validate the StOrM model and one to present it with a more complex dataset. Datasets comprising of $\chi$ and $\chi^*$ needed to be chosen. As $\chi^*$ comprises of expert knowledge which correlates with $\chi$, such expert knowledge had to be found. Behavioural analysis of running processes was chosen as the target domain and discrimination of networking applications as a problem area. Abundant expert knowledge exists in this domain.

**Technical Data - Behaviour of Running Processes :** In order to collect "technical" data, $\chi$, Microsoft Performance Counters API [8] was used. The following seven process-specific attributes and one system wide attribute were selected to be monitored: IO Write Operations/sec, IO Read Operations/sec, IO Other Operations/sec, IO Data Operations/sec, % Privileged Time, % Processor Time, % User Time, Datagrams Sent/sec . This set of eight features yields the ability to observe behaviour of running processes based on their I/O activity, CPU and network usage on the Windows OS. For detailed explanation of each attribute, the reader is referred to [8]. The data was normalised and transformed into an 8-dimensional input feature vector, x.

**TLR Knowledge - Static Analysis of Executables :** As we desire to discriminate between processes that perform networking activity and non-networking applications, suitable expert knowledge from $\chi^*$ had to be chosen. A set of Windows API calls used for network communication in Windows OS was selected from MSDN library [8]. This library is a resource where expert knowledge on various Windows specific libraries is presented and categorised according to various system functions. The following set of strings, representative of numerous API calls, was chosen from a set of libraries that are used for networking within the operating system: Internet, Ftp, Http, WinHttp, WSA, Rpc, Uuid, Dns, Dhcp, Netbios, Net, Snmp, WNet. These strings, which represent more than 90% of Windows networking functions, were selected as our TLR knowledge and encoded in e. Static binary analysis of running processes was then performed, in order to evaluate which API calls each process imports. This information was subsequently transformed into the input feature vector s, representing API calls present in an executable.

### 5.1 Results

**Experiment I -** In the first experiment two running processes were observed for the duration of approximately 150 seconds. Namely the editor Notepad and messaging client MSN Live Messenger. These two applications were chosen due to their difference in terms of networking functionality. In this experiment we want to show that by incorporating extra knowledge as part of the SOM algorithm, we can identify which cluster within the SOM output corresponds to networking activity and thus identify the Messenger application. If we provide any machine learning algorithm with our "technical" data then we can generate a set of clusters, but without labels, and therefore we are unable to determine which cluster belongs to which activity/process. Using our encoded TLR information we can provide enough information in order to help us distinguish between clusters that denote the activity of interest and clusters that are irrelevant to our problem.

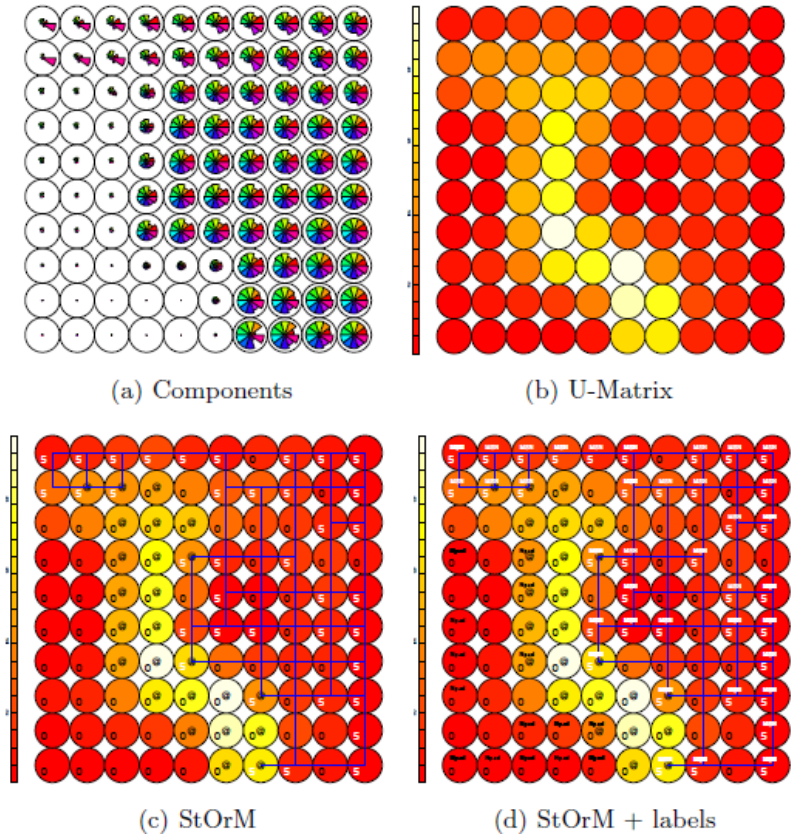

**Fig. 1.** SOM and StOrM results for Experiment I - In (c) and (d) the "@" sign denotes nodes on cluster boundaries, lines connect nodes belonging to cluster of interest and numbers show the amount of TLRs flagged during training.

Results of this experiment can be seen in Figure 1. Figure 1(a) shows component planes [6] of the SOM. This is a standard method for visualising reference vectors of each node in the map. Each component (attribute) is represented as a pie slice, where the size shows the magnitude of this attribute in a given node.

Table 1. Experimental Results

| Experiment | Nodes (i) Total | Winner Nodes ($m_c$) | | | |
|---|---|---|---|---|---|
| | | Quantity | | Mean Net. Level | |
| | | TLR.on | TLR.off | TLR.on | TLR.off |
| I | 100 | 35 | 14 | 0.9873 | 0.0029 |
| II | 400 | 137 | 131 | 0.3151 | 0.1830 |

Figure 1(b) shows the map using the U-Matrix [9] method. These two visualisations are standard methods for presenting the output of SOM. Even though we can see that possibly two clusters exist in our data, it is difficult to distinguish which one corresponds to networking activity and thus Messenger. Without labels and much understanding of the data, such discrimination is difficult. On the other hand with the help of our StOrM model and the incorporation of some expert knowledge, we can automatically delineate the cluster of interest, seen in Figure 1(c). This connected set of nodes highlights a cluster representative of the behaviour of the Messenger application. This result can be validated against a map with labels, showing the labelled true cluster region in Figure 1(d).

The experiment has been run ten times, yielding the following results, seen in Table 1. Out of a total of 100 nodes, on average 35% of nodes were winner nodes, correlated with given expert knowledge from $\chi^*$. Fourteen percent of all nodes were winner nodes that had no correlation at all. In order to confirm the correlation correctness we look at the mean networking level for correlated versus uncorrelated winner nodes. We perform this calculation as we are interested in finding groups or clusters of nodes which represent applications denoting networking behaviour. In this experiment correlated nodes, all identified correctly as belonging to Messenger, have an average networking level of almost 99% of the total networking activity present during the experiment, whereas uncorrelated nodes, all belonging to "Notepad" have on average below 1%.

From the above analysis and the four figures it can be seen that our StOrM model provides a way of correlating expert knowledge with standard "technical" information for the purpose of cluster identification and labelling. This correlation helps with the identification of data or clusters of interest which, without labels, would otherwise be difficult to identify.

Experiment II - In order to assess our StOrM model on a more complex dataset, a larger number of running processes were monitored and subsequently analysed. In total 33 running applications were monitored during a session of standard use of the host machine for approximately 200 seconds.

Figure 2 and Table 1 show results of this experiment. With a more complex dataset it is more difficult to interpret results using the standard techniques seen in Figures 2(a) and 2(b). It is possible to deduce that a number of clusters exist within the underlying data, however which of those are the clusters of interest is very difficult to discern. With the help of our StOrM model however, the un-derstanding of the SOM output becomes much easier, this can be seen in Figure 2(c). Connected regions of the map clearly highlight a cluster of nodes denoting applications that exhibit networking activity due to their implementation of networking functions present in the Windows OS. Labels in Figure 2(d) show that applications that one would intuitively regard as having networking functionality are grouped within the connected region and non-networking applications lie outside of this cluster. This is true for most cases with a handful of exceptions. These are attributed to the fact that such applications either do not use Windows networking functions or have not been active during the session. To validate the correct delineation of the networking cluster, quantitative analysis was again performed on correlated versus uncorrelated node mean networking activity. In this case the discrimination between the two groups is not as apparent as in experiment one. This is due to two reasons. The networking attribute used in our "technical" data is a global measure, rather than a per process signal. Secondly not all applications use Windows networking functions for communication. Again 10 runs have been performed and analysed. The produced SOM output contains 400 nodes, out of which 137 winner nodes have correlated expert knowledge and 131 winner nodes have no correlation. The average networking activity for correlated nodes (32% of total networking activity) is approximately 14% higher than that of uncorrelated nodes (18% of total networking activity). This result again confirms that the cluster highlighted by the StOrM model delineates nodes which are representative of the class of interest.

## 6   Conclusions

In this work we have proposed a novel immune-inspired idea that provides new possibilities for knowledge discovery and exploratory data analysis. The proposed StOrM model incorporates an analogy of the so-called Toll-like receptors from the human immune system. This model provides an insight into a new class of learning where additional knowledge can be fused with traditional "technical" data. This additional information does not need to belong to the same hypothetical space as knowledge encoded within a training or testing dataset. The proposed model is explored with the help of two experiments, grounded within behavioural analysis of running processes on a host system. This experimental domain provides for meaningful selection of expert knowledge encoded within our model. Such knowledge is in the form of expert information as provided by existing functional categorisation of programming methods implemented within the Windows OS. Performed experiments show encouraging results in terms of improved visualisation for exploratory data analysis and knowledge discovery due to the proposed automatic cluster interpretation algorithm.

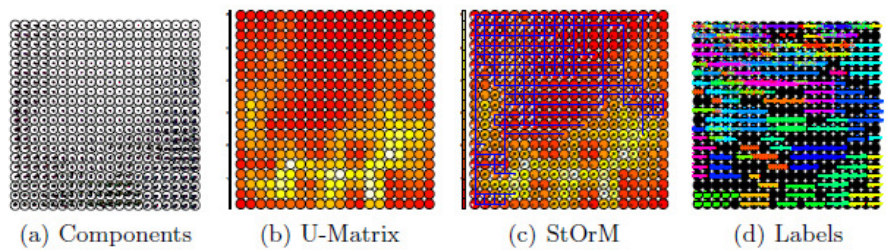

(a) Components    (b) U-Matrix    (c) StOrM    (d) Labels

Fig. 2. SOM and StOrM results for Experiment II

Our model highlights a unique type of learning which becomes especially useful, where no labelled examples exist but some knowledge about classes of interest is acknowledged. From the field of information security all the way to medical sciences, many domains where expert knowledge is abundant exist, yet such knowledge is difficult to incorporate within traditional knowledge discovery techniques. We believe that our technique is a step forward towards combining such disparate knowledge with traditional sources of information and that such fusion can greatly improve understanding of the problem at hand.